\documentclass[final]{cvpr}

\usepackage{times}
\usepackage{epsfig}
\usepackage{graphicx}
\usepackage{amsmath}
\usepackage{amssymb}

\usepackage{multirow}
\usepackage[ruled,vlined]{algorithm2e}
\usepackage{pifont}% http://ctan.org/pkg/pifont
\newcommand{\cmark}{\ding{51}}%
\newcommand{\xmark}{\ding{55}}%

\usepackage{booktabs}  % Professional tables
\usepackage{comment}

\usepackage[colorlinks,bookmarks=false]{hyperref}

\def\cvprPaperID{6104} % *** Enter the CVPR Paper ID here
\def\confYear{CVPR 2021}

\begin{document}

%%%%%%%%% TITLE
\title{Scale-Localized Abstract Reasoning}

\author{
Yaniv Benny\textsuperscript{\rm 1}\\
\and
Niv Pekar\textsuperscript{\rm 1}\\
\and
Lior Wolf\textsuperscript{\rm 1,2}\\
\and
\textsuperscript{\rm 1}The School of Computer Science, Tel Aviv University\\
\textsuperscript{\rm 2}Facebook AI Research (FAIR)\\
}

\maketitle

\begin{abstract}
We consider the abstract relational reasoning task, which is commonly used as an intelligence test. 
Since some patterns have spatial rationales, while others are only semantic, we propose a multi-scale architecture that processes each query in multiple resolutions. We show that indeed different rules are solved by different resolutions and a combined multi-scale approach outperforms the existing state of the art in this task on all benchmarks by 5-54\%.
The success of our method is shown to arise from multiple novelties. First, it searches for relational patterns in multiple resolutions, which allows it to readily detect visual relations, such as location, in higher resolution, while allowing the lower resolution module to focus on semantic relations, such as shape type. Second, we optimize the reasoning network of each resolution proportionally to its performance, hereby we motivate each resolution to specialize on the rules for which it performs better than the others and ignore cases that are already solved by the other resolutions. Third, we propose a new way to pool information along the rows and the columns of the illustration-grid of the query. Our work also analyses the existing benchmarks, demonstrating that the RAVEN dataset selects the negative examples in a way that is easily exploited. We, therefore, propose a modified version of the RAVEN dataset, named RAVEN-FAIR. 
Our code and pretrained models are available at \url{https://github.com/yanivbenny/MRNet}.
\end{abstract}

\section{Introduction}

Raven's Progressive Matrices (RPM) is a widely-used intelligence test~\cite{raven2003raven,carpenter1990one}, which does not require prior knowledge in language, reading, or arithmetics. While IQ measurements are often criticized~\cite{te2001practice,flynn1987massive}, RPM is highly correlated with other intelligence-based properties~\cite{snow1984topography} and has a high statistical reliability~\cite{neisser1996intelligence,mackintosh2011iq}. Its wide acceptance by the psychological community led to an interest in the AI community. Unfortunately, as pointed out by~\cite{szegedy2013intriguing,zhang2016understanding,jo2017measuring}, applying machine learning models to it can sometimes result in shallow heuristics that have little to do with actual intelligence. Therefore, it is necessary to study the pitfalls of RPMs and the protocols to eliminate these exploits.

In Sec.~\ref{sec:datasets}, we present an analysis of the most popular machine learning RPM benchmarks, PGM~\cite{santoro2018measuring} and RAVEN~\cite{zhang2019raven}, from the perspective of biases and exploits. It is shown that RAVEN is built in a way that enables the selection of the correct answer with a high success rate without observing the question itself. To mitigate this bias, we construct, in Sec.~\ref{sec:ravenfair}, a fair variant of the RAVEN benchmark. 
To further mitigate the identified issues, we propose a new evaluation protocol, in which every choice out of the multiple-choice answers is evaluated independently. 
This new evaluation protocol leads to a marked deterioration in the performance of the existing methods and calls for the development of more accurate ones that also allows for a better understating of abstract pattern recognition. In Sec.~\ref{sec:method}, we propose a novel neural architecture, which, as shown in Sec.~\ref{sec:exp}, outperforms the existing methods by a sizable margin. The performance gap is also high when applying the network to rules that were unseen during training. Furthermore, the structure of the new method allows us to separate the rules into families that are based on the scale in which reasoning occurs.

The success of the new method stems mostly from two components: (i) a multi-scale representation, which is shown to lead to a specialization in different aspects of the RPM challenge across the levels, and (ii) a new form of information pooling along the rows and columns of the challenge's $3\times 3$ matrix. 

To summarize, the contributions of this work are as follows: (i) An abstract multi-scale design for relational reasoning. (ii) A novel reasoning network that is applied on each scale to detect relational patterns between the rows and between the columns of the query grid. (iii) An improved loss function that balances between the single positive example and the numerous negative examples. (iv) A multi-head attentive loss function that prioritizes the different resolutions to specialize in solving different rules. (v) A new balanced version of the existing RAVEN dataset, which we call RAVEN-FAIR.

\section{Related Work and Dataset Analysis}\label{sec:related}

The first attempt in an RPM-like challenge with neural networks involved simplified challenges~\cite{hoshen2017iq}. The Wild Relation Network (WReN)~\cite{santoro2018measuring}, which relies on the Relation Module~\cite{santoro2017simple}, was the first to address the full task and introduced the PGM benchmark. 
WReN considers the possible choices one-by-one and selects the most likely option.
Two concurrently proposed methods, CoPINet~\cite{zhang2019learning} and LEN~\cite{zheng2019abstract} have proposed to apply row-wise and column-wise relation operations. Also, both methods benefit from processing all eight possible choices at once, which can improve the reasoning capability of the model, but as we show in this work, has the potential to exploit biases in the data that can arise during the creation of the negative choices. CoPINet applies a contrast module, which subtracts a common factor from all choices, thereby highlighting the difference between the options. LEN uses an additional ``global encoder'', which encodes both the context and choice images into a single global vector that is concatenated to the row-wise and column-wise representations. The latest methods, MXGNet~\cite{wang2020abstract} and Rel-AIR~\cite{spratleycloser}, have proposed different complex architectures to solve this task and also consider all choices at once.

Zhang et al.~\cite{zheng2019abstract} also introduced a teacher-student training method. It selects samples in a specialized category- and difficulty-based training trajectory and improves performance. In a different line of work, variational autoencoders~\cite{kingma2013auto} were shown to disentangle the representations and improve generalization on held-out rules~\cite{steenbrugge2018improving}. Our method shows excellent performance and better generalization to held-out rules without relying on either techniques.

When merging different outputs of intermediate paths, such as in residual~\cite{he2016deep} or shortcut~\cite{bishop1995neural,ripley2007pattern,venables2013modern} connections, most methods concatenate or sum the vectors into one. This type of pooling is used by the existing neural network RPM methods when pooling information across the grid of the challenge. In Siamese Networks~\cite{bromley1994signature}, one compares the outputs of two replicas of the same network, by applying a distance measure to their outputs. Since the pooling of the rows and the columns is more akin to the task in siamese networks, our method generalizes this to perform a triple pair-wise pooling of the three rows and the three columns. 

\begin{figure}[t]
\centering
\begin{tabular}{c@{~}c}
\includegraphics[width=0.22\textwidth, trim={3 4 4 2}, clip]{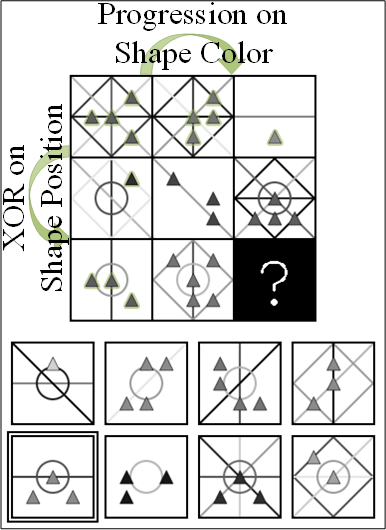} &
\includegraphics[width=0.22\textwidth, trim={3 4 4 2}, clip]{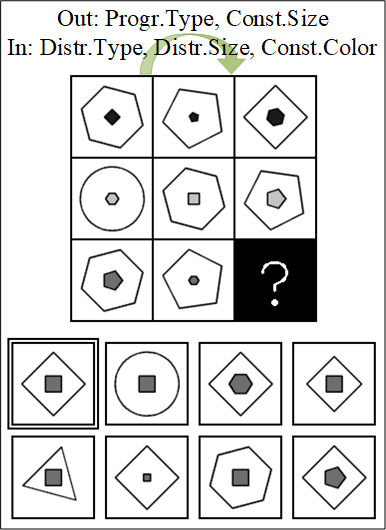} \\
(a)&(b) \\
\end{tabular}
\caption{Dataset examples. (a) PGM. (b) RAVEN. The rules are annotated and the correct answer is highlighted.}
\label{fig:dataset}
\end{figure}

\subsection{Datasets}\label{sec:datasets}

The increased interest in the abstract reasoning challenges was enabled by the introduction of the PGM~\cite{santoro2018measuring} and RAVEN~\cite{zhang2019raven} datasets. 
These datasets share the same overall structure. The participant is presented with the first eight images of a 3x3 grid of images, called the context images, and another eight images, called the choice images. The objective is to choose the missing ninth image of the grid out of the eight presented choices, by identifying one or more recurring patterns along the rows and/or columns of the grid. The correct answer is the one that fits the most patterns. See Fig.~\ref{fig:dataset} for illustration. 

PGM is a large-scale dataset consisting of 1.2M train, 20K validation, and 200K test questions. All images in the dataset are conceptually similar, each showing various amounts of lines and shapes of different types, colors, and sizes. Each question has between 1-4 rules along the rows or columns. Each applies to either the lines or the shapes in the image. Fig.~\ref{fig:dataset}(a) shows an example of the dataset. 

RAVEN is a smaller dataset, consisting of 42K train, 14K validation, and 14K test questions divided into 7 distinct categories. Each question has between 4-8 rules along the rows only. Fig.~\ref{fig:dataset}(b) shows an example of the dataset.

\paragraph{Evaluation protocols}
Both datasets are constructed as a closed-ended test. It can be performed as either a multiple-choice test (MC) with eight choices, where the model can compare the choices, or as a single choice test (SC), where the model scores each choice independently and the choice with the highest score is taken. While this distinction was not made before, the previous works are divided in their approach. WReN follows the SC protocol and CoPINet, LEN, MXGNet, and Rel-AIR, all followed the MC protocol.

In the SC scenario, instead of presenting the agent with all the choices at once, it is presented with a single choice and has to predict if it is the right answer or not. The constraint of solving each image separately increases the difficulty since the model cannot make a decision by comparing the choices. While this eliminates the inter-choice biases, the agent may label multiple or zero images as the correct answer and opens the door to multiple success metrics. In order to directly compare models trained in the MC and SC protocols, we evaluate both types of models in a uniform manner: the score for all models is the accuracy for the multiple-choice test, where for the SC-trained model, we consider the answer with the highest confidence, regardless of the number of positive answers.

\paragraph{Dataset Analysis} 
When constructing the dataset for either the MC or the SC challenge, one needs to be very careful in how the negative examples are selected. If the negative examples are too obvious, the model can eliminate them and increase the probability of selecting the correct answer, without having to fully identify the underlying pattern. Negative examples are therefore constructed based on the question.
However, when the negative examples are all conditioned on the correct answer, in the MC scenario, the agent might be able to retrieve the correct answer by looking at the choices without considering the context at all. For example, one can select the answer that has the most common properties with the other answers. See the supplementary for an illustration of such biases in a simple language-based multiple-choice test.

A context-blind test, which is conducted by training a model that does not observe the context images, can check whether the negative answers reveal the correct answer. Ideally, the blind test should return a uniform random accuracy, e.g. 12.5\% for eight options.
However, since the negative choices should not be completely random so that they will still form a challenge when the context is added, slightly higher accuracy is acceptable. When introduced, the PGM dataset achieved a blind test score of 22.4\%. In the RAVEN benchmark, the negative examples are generated by changing a single attribute from the correct image for each example, making it susceptible to a majority-based decision. RAVEN was released without such a context-blind test, which we show in the supplementary that it fails at.

Concurrently to our work, Hu et al.~\cite{hu2021stratified} have also discovered the context-blind flaw of RAVEN. They propose an adjustment to the dataset generation scheme that eliminates this problem, which they call `Impartial-RAVEN'. Instead of generating the negative choices by changing a random attribute from the correct answer, they propose to sample in advance three independent attribute changes and generate seven images from all the possible combinations of them. 

\section{RAVEN-FAIR}
\label{sec:ravenfair}
In the supplementary, we analyze both PGM and RAVEN under the blind, SC, and MC settings. As we show, due to the biased selection of the negative examples, the RAVEN dataset fails the context-blind test, as it is solved with 80.17\% accuracy by only looking at the choices, making it unsuitable for the MC protocol.

We, therefore, propose a modified dataset we term RAVEN-FAIR, generated by Algorithm.~\ref{algorithm1}. This algorithm starts with a set of choices that contains the correct answer and iteratively enlarges this set, by generating one negative example at a time. At each round, one existing choice (either the correct one or an already generated negative one) is selected and a new choice is created by changing one of its attributes. The result is a connected graph of choices, where the edges represent an attribute change.

\begin{algorithm}[t]
    \SetAlgoLined
    \SetKwInOut{Input}{input}
    \SetKwInOut{Output}{output}
    \Input{$C$ - 8 context components \\ $a$ - the correct answer for $C$}
    \Output{$A$ - 8 choice components ($a \in A$)}
     $A \xleftarrow{} \{ a \}$\;
     \While{length(A) $<$ 8}{
      $a' \xleftarrow{} Choice(A)$\;
      $\hat{a} \xleftarrow{} Modify(a')$\;
      \If{Solve($C,\hat{a}$) = False}{
       $A \xleftarrow{} A \cup \{\hat{a}\}$\;
      }
     }
    \caption{RAVEN-FAIR}\label{algorithm1}
\end{algorithm}

\begin{figure}[t]
\centering
    \begin{tabular}{cc}
    \includegraphics[width=0.45\columnwidth]{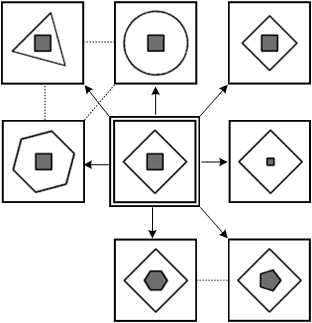} &
    \includegraphics[width=0.45\columnwidth]{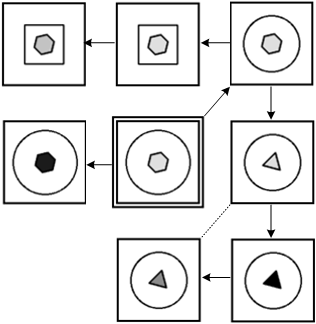} \\
    (a) & (b) \\
    \end{tabular}
    \caption{Illustration of negative examples' generation processes. (a) RAVEN. (b) RAVEN-FAIR.}
    \label{fig:algorithm_old}
\end{figure}

As an example, we show the process of generating the negative examples for RAVEN in Fig.~\ref{fig:algorithm_old}(a). For comparison, we also show the process of how our proposed algorithm generates the negative examples for RAVEN-FAIR in Fig.~\ref{fig:algorithm_old}(b). Note that the two figures show the produced choices for different context questions. In each figure, eight choices for an initial question are shown. The center image, which is also highlighted, is the correct answer. Each arrow represents a newly generated negative example based on an already existing choice, by changing one arbitrary attribute. The dotted lines connect between choices that also differ by one attribute but were not generated with condition to each other.
As can be seen, the RAVEN dataset generates the questions in a way that the correct answer is always the one with the most shared attributes with the rest of the examples. The negative option with the most neighbors in this example is the top-left one, which only has three neighbors, while the correct answer always has eight. Even without highlighting the correct answer, it would be easy to point out which one it is, by selecting the one with the most neighbors, without looking at the question. 

In contrast, our fair algorithm generates the negative examples in a more balanced way. Since each newly generated negative choice can now be conditioned on both the correct image or an already existing negative one, the correct choice cannot be tracked back by looking at the graph alone. In this example, the correct answer only has two neighbors, and both the right and top-right negative images have three neighbors. Because the algorithm is random, the number of neighbors that the correct image has is arbitrary across the dataset, and is between one and eight.

In the context-blind test (supplementary), RAVEN-FAIR returned a 17.24\% accuracy, therefore passing the test and making it suitable for both SC and MC. The PGM dataset passed our context-blind test as well, with 18.64\% accuracy.

\section{Method}\label{sec:method}

\begin{figure*}[t]
\centering
\includegraphics[width=0.9\textwidth, trim={5 10 5 10}, clip]{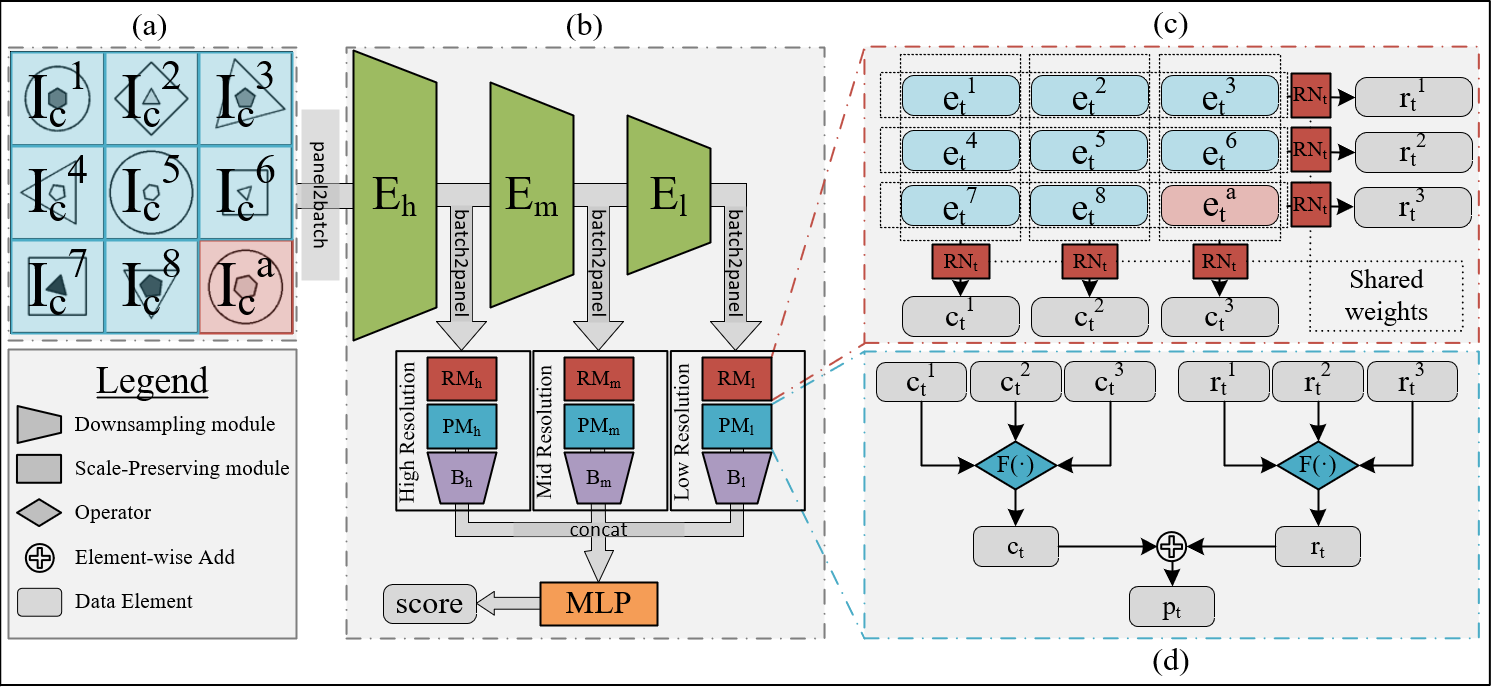}
\caption{MRNet. (a) Depiction of the input query. (b) High-level diagram of the architecture. (c) The relation module ($RM_t$). (d) The pattern module ($PM_t$).}
\label{fig:architecture}
\end{figure*}

Our Multi-scale Relation Network (MRNet), depicted in Fig.~\ref{fig:architecture}(b), consists of five sub-modules: (i) a three-stage encoder $E_t$, where $t\in\{h,m,l\}$, which codes the input context and a single choice image into representations in three different resolutions: $20\times 20$ (high), $5\times 5$ (med) and $1\times 1$ (low), (ii) three relation modules $RM_t$, one for each resolution, which perform row-wise and column-wise operations on the encodings to detect relational patterns, (iii) three pattern modules $PM_t$, one for each resolution, which detects if similar patterns occur in all rows or columns, (iv) three bottleneck networks $B_t$, which merge the final features of each resolution, and (v) a predictor module $MLP$, which estimates the correctness of a given choice image to the context in question, based on the bottlenecks' outputs. 
The model is presented with a question in the form of $16$ images: a set of context images $I_C = \{I^n | n \in [1,8]\}$ and a set of choice images $I_A = \{I_a^n | n \in [1,8]\}$. 
Since the model operates in the SC protocol, it evaluates the choices separately. Therefore, the notations act as if there is only a single choice image $I^a$ to answer. To solve all eight choices, the model is repeated eight times with a different choice image $I_a\in I_A$.

\paragraph{Multi-scale encoder}
The encoder is a three-stage Convolutional Neural Network that encodes all images (context or choice) into multi-resolution representations (high, middle, low). Every image $I^n \in [-1,1]^{1,80,80}$, for $n \in \{1,2,...,8,a\}$, is passed sequentially through the three stages of the encoder, i.e., the middle  resolution encoding is obtained by applying further processing to the output of the high resolution encoder and similarly the low resolution encoding is obtained by further processing the middle resolution one. Specifically, the encoding process results in three tensors. $e_h^n \in \mathbb{R}^{64,20,20}$, $e_m^n \in \mathbb{R}^{128,5,5}$ and $e_l^n \in \mathbb{R}^{256,1,1}$. 
\begin{equation}\label{eq:encoder}
    e^n_h = E_h(I^n), \quad e_m^n = E_m(e_h^n), \quad  e_l^n = E_l(e_m^n)\,.
\end{equation}

\paragraph{Relation Module}
Based on the encoding of the images, we apply a relational module (RM) to detect patterns on each row and column of the query. There are three such modules ($RM_h, RM_m, RM_l$). For each resolution $t \in \{h,m,l\}$, the 9 encodings $e^n_t$ for $n=1,2,..,8,a$ are positioned on a 3x3 grid, according to the underlying image $I^n$, see Fig.~\ref{fig:architecture}(c). 
The rows and columns are combined by concatenating three encodings on the channel dimension. The rows consist of the three triplets $(e^1_t,e^2_t,e^3_t)$, $(e^4_t,e^5_t,e^6_t)$, $(e^7_t,e^8_t,e^a_t)$. Similarly, the columns consist of the three triplets $(e^1_t,e^4_t,e^7_t)$, $(e^2_t,e^5_t,e^8_t)$, $(e^3_t,e^6_t,e^a_t)$. Each row and column is passed through the relation network (RN).
\begin{equation}\label{eq:rm}
\begin{split}
    r^1_t = RN_t(e^1_t,e^2_t,e^3_t), \quad c^1_t = RN_t(e^1_t,e^4_t,e^7_t) \\
    r^2_t = RN_t(e^4_t,e^5_t,e^6_t), \quad c^2_t = RN_t(e^2_t,e^5_t,e^8_t) \\
    r^3_t = RN_t(e^7_t,e^8_t,e^a_t), \quad  c^3_t = RN_t(e^3_t,e^6_t,e^a_t)
\end{split}
\end{equation}
Each RN consists of two residual blocks with two convolution layers inside each one. 
For the high and middle resolutions, the convolution has a kernel of size 3 with 'same' padding, while for the low resolution the kernel size is 1 without padding. 
The output of each relation block is of the same shape as its corresponding encoding $e_t$, i.e. $RN_h: \mathbb{R}^{3*64,20,20} \rightarrow \mathbb{R}^{64,20,20}$, $RN_m: \mathbb{R}^{3*128,5,5} \rightarrow \mathbb{R}^{128,5,5}$, and $RN_l: \mathbb{R}^{3*256,1,1} \rightarrow \mathbb{R}^{256,1,1}$.
Note that we apply the same relation networks to all rows and columns. This approach allows comparison between the rows and between the columns to detect recurring patterns. In addition, it maintains the permutation and transpose invariance property we assume on the rows and columns.

\paragraph{Pattern Module}
At this point, at each level $t \in \{h,m,l\}$ the representation of the panel is structured as three row features and three column features, which we want to merge into a single representation. The module dedicated to this is the pattern module (PM), which applies some operator $F(\cdot)$ on the rows and columns. It is depicted in Fig.~\ref{fig:architecture}(d). 

To promote order-invariance between the three rows or columns, a permutation-invariant operator $F(\cdot)$ is recommended. One can perform sum pooling (SUM3). This kind of approach was employed by~\cite{santoro2018measuring,zhang2019learning,zheng2019abstract}. Due to the linearity of the sum pooling, a ReLU is employed.
\begin{equation}\label{eq:defsum}
\text{SUM3}(x_1,x_2,x_3) := \text{ReLU}(x_1 + x_2 + x_3)
\end{equation}
The reduction is then applied on the rows and columns.
\begin{equation}\label{eq:sum}
    r_t = \text{SUM3}(r^1_t, r^2_t, r^3_t), \quad 
    c_t = \text{SUM3}(c^1_t, c^2_t, c^3_t)
\end{equation}
The SUM3 operator has a drawback in that it does not detect similarity between the rows and between the columns. 
Instead, we propose a novel method that is inspired by Siamese Networks, which compares multiple rows and columns.
This vector operator DIST3 is defined per vector index $i$ and is applied to the rows and columns:
\begin{equation}\label{eq:defdist}
\begin{split}
\text{DIST3}_i(x_1,x_2,x_3) :=& (x_{1,i} - x_{2,i})^2 \\
    +& (x_{2,i} - x_{3,i})^2 \\
    +& (x_{3,i} - x_{1,i})^2
\end{split}
\end{equation}
\begin{equation}\label{eq:dist}
\begin{gathered}
    r_t = \text{DIST3}(r^1_t, r^2_t, r^3_t),\quad
    c_t = \text{DIST3}(c^1_t, c^2_t, c^3_t) \\
\end{gathered}
\end{equation}
DIST3 does not contain an activation function, since the method is already non-linear. Its  advantage is also visible in its backward path. The gradient of each operator with respect to element $i$ of $x_1$ is:
\begin{equation}\label{eq:defgrad}
\begin{split}
\frac{\partial\text{SUM3}_i(x_1,x_2,x_3)}{\partial x_{1,i}} &= 1  \\
\frac{\partial\text{DIST3}_i(x_1,x_2,x_3)}{\partial x_{1,i}} &= 2\cdot(2\cdot x_{1,i} - x_{2,i} -x_{3,i}) 
\end{split}
\end{equation}
The gradient of SUM3 does not depend on the values of $x_2$ and $x_3$, while the gradient of DIST3 does. Therefore, DIST3 has the potential to encourage increased coordination between the paths. 

Finally, the row- and column-features are summed into a merged representation of the entire panel:
\begin{equation}
p_t = r_t + c_t
\end{equation}

\paragraph{Bottleneck module}
The three relation modules and their pattern detectors return a latent representation of the panel in three resolutions ($p_h, p_m, p_l$). This module collects the three representations into a single representation. For this purpose, each representation is downsampled by a bottleneck network ($B_t$) that encodes the panel representation $p_t$ into feature vectors with 128 features.
\begin{equation}
\label{eq:bottleneck}
v_t = B_t(p_t)
\end{equation}
The three final features are then concatenated into a single vector with 384 features {(3x128)}.
\begin{equation}
v = \text{Concat}(v_h, v_m, v_l)
\end{equation}

\paragraph{The predictor}
The resulting merged feature vector from the bottleneck module is used to predict the correctness of a choice $I^a \in I_A$ to the context $I_C$. A Multi-Layer Perceptron (MLP) predicts the score of the choice images. A sigmoid translates this score to the probability of the choice $I^a$ to be a correct answer:
\begin{equation}\label{eq:mlp}
p(y=1|I^a,I_C) = \text{Sigmoid}(MLP(v))
\end{equation}
For each choice image, the loss is the Binary Cross Entropy loss. Because each grid has seven negative choices and only one correct answer, we weight the loss of each choice respectively, meaning that the correct answer has a weight of 7 and each negative answer has a weight of 1.

\paragraph{Multi-head attentive predictor}
Optimizing the three heads with the main loss works relatively well, but we have found that adding an objective that optimizes each head separately improves the final results and reduces the overall training time. To do so, three additional predictors were created, one for each bottleneck output $v_h,v_m,v_l$. The architecture of the heads is the same as that of the main head and the loss function is identical. To encourage each resolution to specialize in the rules that the other resolutions are having difficulty with, we propose an attentive weight distribution on the loss functions of the three heads. Consider the prediction of each head $p_t(y=y^*|I^a,I_C)$ for $t \in \{h,m,l\}$, where $y^* \in \{0,1\}$ represent whether the choice $I^a$ is correct or not. The  weight applied to each head is:
\begin{equation}
    w_t = \frac{\exp\{p_t(y=y^*|I^a,I_C)\}}{\sum_{\tau}\exp\{p_\tau(y=y^*|I^a,I_C)\}}
\end{equation}

Let $\mathcal{L}_t$ be the Binary Cross Entropy loss for head $t \in \{h,m,l\}$. The final multi-head loss is:
\begin{equation}
    \mathcal{L}_3 = \sum_t w_t \mathcal{L}_t
\end{equation}

\noindent{\bf Training method\quad}
We trained with a batch size of 32 queries with an early stopping scheme of 20 epochs, measuring the accuracy on the validation set. The reported accuracy was measured on the test set, for the checkpoint with the best validation accuracy. See supplementary for a detailed description of the architecture. We did not tune any hyper-parameter of the model. Adam~\cite{kingma2014adam} was used with lr=1e-3, $\beta$=(0.9,0.999), and weight decay of 1e-6.

\begin{table}[t]
\centering
\begin{tabular}{@{}l@{~}l@{~}l@{}c@{~}c@{~}c@{~}c@{~}c@{}}
\toprule
& \multicolumn{7}{c}{Test Accuracy (\%)} \\
\midrule
& \multicolumn{3}{c}{Model} & \multicolumn{2}{c}{PGM} & \multicolumn{2}{c}{RAVEN} \\
\cmidrule(lr){2-4}
\cmidrule(lr){5-6}
\cmidrule(lr){7-8}
& Name & Version & SC & base & meta & FAIR & orig \\

\midrule
\multirow{5}{*}{\rotatebox[origin=c]{90}{SC Baselines}}
% SC BAsELINES
& \emph{ResNet-SC}           &               &  \cmark   & $^\dagger$48.9 & -  & $^\dagger$58.3 & $^\dagger$40.4 \\
& \emph{WReN'18}%\comment{\cite{santoro2018measuring}} 
                            & normal        &  \cmark   & 62.6 & 76.9 & $^\dagger$30.3 & $^\dagger$16.8 \\
& \emph{V-WReN'18}%\cite{steenbrugge2018improving} 
                            & Normal        &  \cmark   & 64.8 & -  & - & - \\
& \emph{CoPINet'19}%\shortcite{zhang2019learning} 
                            & no-contrast   &  \cmark   & $^\dagger$42.7 & $^\dagger$45.2     & $^\dagger$36.5 & $^\dagger$18.4 \\
& \emph{LEN'19}%\shortcite{zheng2019abstract}     
                            & global-8      &  \cmark   & $^\dagger$65.6 & $^\dagger$79.6 & $^\dagger$50.9 & $^\dagger$29.9 \\
\midrule
& \emph{MRNet}              &           &  \cmark & {\bf93.4} & {\bf92.6} &  {\bf86.8} & {\bf84.0} \\
\cmidrule(lr){1-8}
\multirow{5}{*}{\rotatebox[origin=c]{90}{Ablation}} 
& \multicolumn{2}{l}{no $\mathcal{L}_{3}$}          &  \cmark & 84.4 & 88.4 &  84.0 & 80.1 \\
& \multicolumn{2}{l}{no $\mathcal{L}_{3}$, no wb}   &  \cmark & 75.2 & 84.9 &  80.6 & 78.6 \\
& \multicolumn{2}{l}{no $\mathcal{L}_{3}$, no wb, with SUM3}  &  \cmark & 74.3 & 79.0 &  77.0 & 69.6 \\
& \multicolumn{2}{l}{no wb}  &  \cmark & 87.6 & 88.9 &  83.4 & 80.2 \\
& \multicolumn{2}{l}{with SUM3}  &  \cmark & 83.2 & 85.3 &  79.5 & 78.2 \\
\midrule
\midrule
\multirow{8}{*}{\rotatebox[origin=c]{90}{MC Baselines}}
% MC BAsELINES
& \emph{ResNet-MC}           &               &  \xmark   & $^\dagger$41.1 & -   & $^\dagger$24.5 & $^\dagger$72.5 \\
& \emph{CoPINet'19}%\shortcite{zhang2019learning} 
                            & normal        &  \xmark   & 56.4 & $^\dagger$51.1 & $^\dagger$50.6 & 91.4 \\
& \emph{LEN'19}%\shortcite{zheng2019abstract}     
                            & normal        &  \xmark   & 68.1 & 82.3 & $^\dagger$51.0  & 72.9 \\
& \emph{LEN'19}%\cite{zheng2019abstract}     
                            & teacher-model &  \xmark   & 79.8 & 85.8 & - & 78.3 \\
& \emph{T-LEN'19}%\cite{zheng2019abstract}     
                            & normal        &  \xmark   & 70.3 & 84.1 & \xmark & \xmark \\
& \emph{T-LEN'19}%\cite{zheng2019abstract}     
                            & teacher-model &  \xmark   & 79.8 & 88.9 & \xmark & \xmark \\
& \emph{MXGNet'20}%\cite{wang2020abstract}   
                            &               &  \xmark   & 66.7 & 89.6 & - & 83.9 \\
& \emph{Rel-AIR'20}%\cite{spratleycloser}   
                            &               &  \xmark   & 85.5 & - & - & 94.1 \\                            
\midrule
& \emph{MRNet}              & with-contrast &  \xmark & {\bf94.5} & {\bf92.8} & {\bf 88.4} & {\bf96.6} \\
\cmidrule(lr){1-8}
\multirow{3}{*}{\rotatebox[origin=c]{90}{Ablation}}
& \multicolumn{2}{l}{no $\mathcal{L}_{3}$}          &  \xmark & 85.7 & 89.0 &  85.2 & 95.5 \\
& \multicolumn{2}{l}{no $\mathcal{L}_{3}$, no wb}   &  \xmark & 76.4 & 85.4 & 81.3 & 94.3 \\
& \multicolumn{2}{l}{no wb}   &  \xmark & 87.4 & 89.8 & 86.1 & 95.0 \\
\bottomrule
\end{tabular}
\caption{Evaluation on all datasets. $^\dagger$Baseline was run by us, due to missing results.}\label{tab:evaluation}
\end{table}

\begin{table}[t]
\centering
\centering
\begin{tabular}{l@{~}c@{~~}c@{~~}cc@{~~}c}
\toprule
& \multicolumn{5}{c}{Accuracy (\%)} \\
\cmidrule(lr){2-6}
 & \multicolumn{3}{c}{Baselines} & \multicolumn{2}{c}{\emph{MRNet}} \\
 \cmidrule(lr){2-4}
 \cmidrule(lr){5-6}
Regime & \emph{WReN} & \emph{V-WReN} & \emph{MXGNet} & (a) & (b) \\
\midrule
Neutral             & 62.6 & 64.2 & 66.7 & {\bf75.2} & {\bf93.4} \\
Interpolation       & 64.4 & -    & 65.4 & {\bf67.1} & {\bf68.1} \\
Extrapolation       & 17.2 & -    & 18.9 & {\bf19.0} & {\bf19.2} \\
HO Pairs            & 27.2 & 36.8 & 33.6 & {\bf37.8} & {\bf38.4} \\
HO Triple Pairs     & 41.9 & 43.6 & 43.3 & {\bf53.4} & {\bf55.3} \\
HO Triples          & 19.0 & 24.6 & 19.9 & {\bf25.3} & {\bf25.9} \\
HO line-type        & 14.4 & -    & 16.7 & {\bf27.0} & {\bf30.1} \\
HO shape-color      & 12.5 & -    & 16.6 & {\bf16.9} & {\bf16.9} \\
\bottomrule
\end{tabular}
\caption{Generalization evaluation on the held-out regimes of PGM. (a) Without weight balancing and $\mathcal{L}_3$. (b) With them.}
\label{tab:generalization}
\end{table}

\begin{table}[t]
\centering
\begin{tabular}{c@{~~}c@{~}cc@{~}c@{~}c@{~}c@{~}c@{~}c@{~}c@{~}c}
\toprule
\multicolumn{3}{c}{{Scales}} & \multicolumn{8}{c}{Configurations. Accuracy(\%)} \\
\cmidrule(lr){1-3}
\cmidrule(lr){4-11}
h & m & l                   & {All} & {Center} & {2$\times$2} & {3$\times$3} & {L-R} & {U-D} & {O-IC} & {O-IG} \\
\midrule
\cmark & \xmark & \xmark    & 72.0 & 84.1 & 47.2 & 48.3 & {\bf90.0} & {\bf91.5} & {\bf87.6} & 55.5 \\ % 100
\xmark & \cmark & \xmark    & {\bf73.3} & 90.5 & 56.1 & 53.4 & 83.5 & 83.0 & 82.5 & {\bf64.1} \\ % 010
\xmark & \xmark & \cmark    & 58.9 & {\bf97.7} & {\bf62.2} & {\bf62.8} & 42.3 & 42.5 & 56.8 & 48.1 \\ % 001
\midrule
\cmark & \cmark & \xmark    & 77.7     & 82.8     & 61.7     & 59.5     & {\bf91.7}& {\bf92.2}& {\bf89.9}& {\bf66.3} \\ % 001
\cmark & \xmark & \cmark    & {\bf78.3}& 91.0     & 63.2     & 64.7     & 88.8     & 92.2     & 87.7     & 60.3 \\ % 001
\xmark & \cmark & \cmark    & 76.6     & {\bf98.3}& {\bf67.5}& {\bf65.4}& 81.6     & 83.4     & 81.5     & 58.3\\ % 001
\midrule
\cmark & \cmark & \cmark    & 80.6 & 87.7 & 64.7 & 64.3 & 94.2 & 94.3 & 90.9 & 68.0 \\ % 111
\multicolumn{3}{c}{{with $\mathcal{L}_3$}} & {\bf86.8} & {\bf97.0} & {\bf72.7} & {\bf69.5} & {\bf98.7} & {\bf98.9} & {\bf97.6} & {\bf73.3} \\ % 111
\bottomrule
\end{tabular}
\caption{Ablation on the role of each scale in RAVEN-FAIR}
\label{tab:multires}
\end{table}

\begin{figure*}[t]
\centering
    \begin{tabular}{@{}c@{}c@{}c@{}c@{}}
    \includegraphics[width=0.091\linewidth, trim={8 5 8 8}, clip]{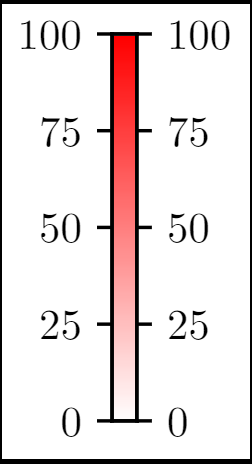} &
    \includegraphics[width=0.293\linewidth, trim={8 5 8 8}, clip]{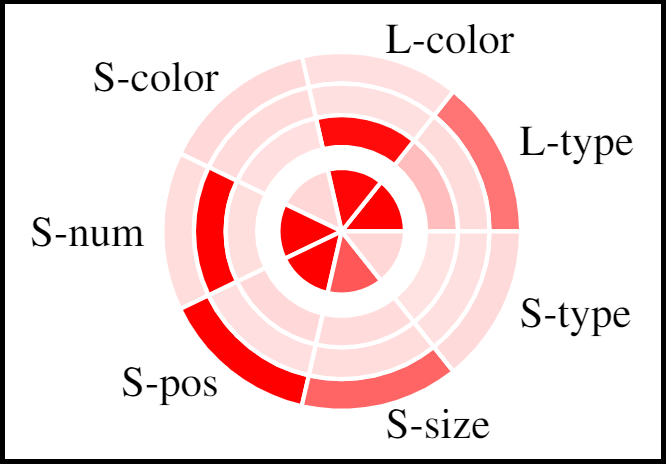} &
    \includegraphics[width=0.293\linewidth, trim={8 5 8 8}, clip]{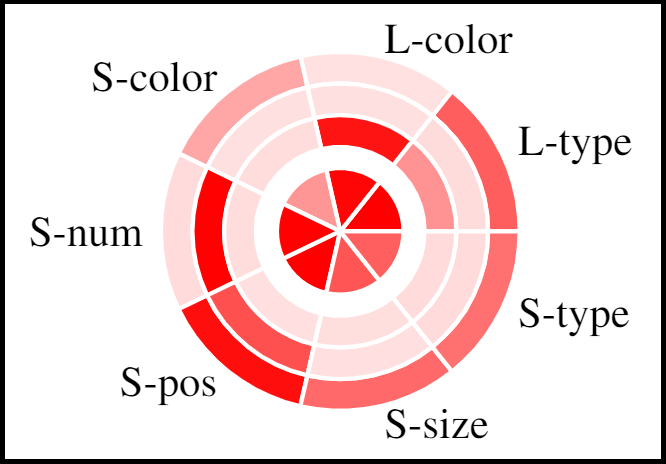} &
    \includegraphics[width=0.293\linewidth, trim={8 5 8 8}, clip]{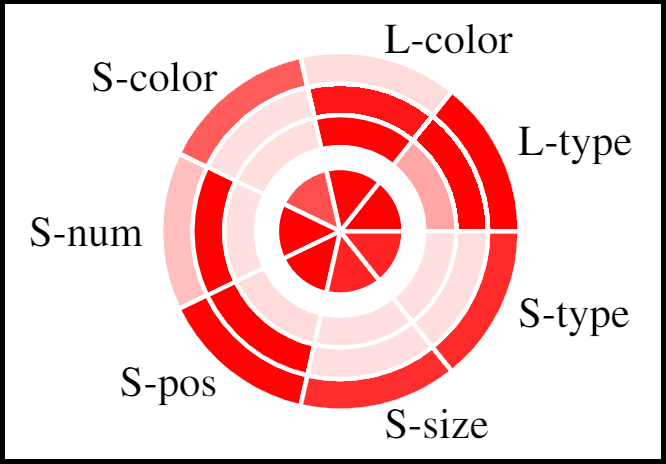} \\
    &(a)&(b)&(c) \\
    \end{tabular}
    \caption{Ablation on the role of each scale in PGM, using a fully trained MRNet on the SC protocol. (a) Without auxiliary loss (PGM), no $\mathcal{L}_3$. (b) With auxiliary loss (PGM\_meta), no $\mathcal{L}_3$. (c) Wihout auxiliary loss (PGM), with $\mathcal{L}_3$. We measured the accuracy for each type, using only one of the scales and the other ones are masked. The inner radius shows the accuracy of the full model. The three rings show the accuracy of each resolution. The inner ring is the low resolution, then the middle, and the high resolution is the outer one.}\label{fig:multires}
\end{figure*}

\section{Experiments}\label{sec:exp}

We experiment on PGM, RAVEN, and RAVEN-FAIR. Each dataset has a preset train, validation, and test splits of the data. We used the train set for training and the test set for the evaluation. The validation set was used to pick the best checkpoint during training for evaluation. PGM has additional metadata of the rules logic, which can be used as additional supervision during training. The use of the metadata has been shown to improve performance on all previous works. We refer to this more informative setting as PGM\_meta. Following previous works, the utilization of the metadata was done by extending the predictor module with an additional head that predicts the 12 bits of the metadata, and training the model with an additional (auxiliary) Cross Entropy loss. Similar to the previous works, the weight for the auxiliary loss is $\beta=10$.

\smallskip
\noindent{\bf Baselines\quad}
We compare our method to the state of the art methods: WReN~\cite{santoro2018measuring}, LEN~\cite{zheng2019abstract}, CoPINet~\cite{zhang2019learning}, MXGNet~\cite{wang2020abstract}, and Rel-AIR~\cite{spratleycloser}. We also employed the ResNet~\cite{he2016deep} models described in the supplementary: ResNet-SC and ResNet-MC. The former evaluates each choice separately and the latter all choices at once.

As noted in Sec.~\ref{sec:related}, LEN and CoPINet follow the MC protocol. To evaluate these baselines without the possibility of exploiting the weaknesses of RAVEN, we have created additional versions of them for the SC protocol.
In LEN, the `global-CNN' originally accepted all 16 images, including the choices. We have changed it to accept only the context images (`global-8'). In CoPINet, we removed the contrasting module (`no-contrast'), which allows information to pass between the choices. 
We also compared to LEN with their proposed teacher-model training schedule, and to T-LEN, which has dedicated prediction paths for line rules and shape rules and is only applicable to PGM.
The results can be seen in Tab.~\ref{tab:evaluation} separately for each protocol.

\smallskip
\noindent{\bf Results\quad}
Before evaluating our model, two important things can be noticed by observing at the baselines. 
First, the performance of LEN and CoPINet declines in SC, where unlike MC, the models cannot compare the choices within the model. This is especially noticeable in RAVEN, where CoPINet practically solved it in MC and failed it in SC. On both RAVEN and RAVEN-FAIR, all SC baselines performed worse than the simple ResNet-SC, which does not have any dedicated modules for relational reasoning. 
Second, by comparing each models' performance between RAVEN and RAVEN-FAIR, we can see that MC models perform significantly lower on RAVEN-FAIR than on RAVEN. This confirms that RAVEN-FAIR indeed fixes the biases in the data and is much more suitable as a benchmark than RAVEN, especially for the MC protocol. 
These observations align with the dataset analysis presented in the supplementary, which conclusions were discussed in Sec.~\ref{sec:datasets}.

One can observe that our model outperforms the baselines in both protocols and across all datasets. In SC, our method outperforms previous baselines by 27.8\% in PGM, 13.0\% in PGM\_meta, 35.9\% in RAVEN-FAIR, and 54.1\% in RAVEN. In MC, we outperform the baselines by 9.0\%, 3.2\%, 37.4\% and 4.9\% respectively. Except for RAVEN, our SC protocol model even performed better than both SC and MC protocols of the baselines, despite the SC protocol having less information. We noticed that our model's performance has reached a point where the auxiliary task of the metadata hurts performance since the added task creates a burden on the optimization of the main task (Vapnik et al. (1998)~\cite{Vapnik1998}) and does no longer benefit the model. Therefore the accuracy on PGM\_meta is lower than that of PGM, which is the first time a model shows such behavior.

For {\bf ablation}, we trained our model with the SUM3 operator instead of DIST3, without the weight balancing of the positive choices versus the negative ones (no wb), and without the multi-head attentive loss ($\mathcal{L}_3$). The results are shown in Tab.~\ref{tab:evaluation} below those of the full model. For SC, we show an aggregated ablation on these modifications, i.e., each following instance applies an additional modification. Each added modification results in a performance decline compared to the previous version. The experiments which remove one component at a time also support that both multi-head attentive loss and the weight balancing both greatly improve the training of the main head. It is also evident that DIST3 is a superior pooling operator over SUM3.  For MC, we performed an ablation with either or both the weight balancing and the $\mathcal{L}_3$ removed. The simplified models without these contributions are not competitive.

Aside from test accuracy, model capacity and training resource consumption is also an important factor. While our method outperforms the other baselines, it is not bigger or slower to train. Our method trains in about 40 minutes per epoch on PGM on a single 1080TI GPU, which is about the same speed as WReN and CoPINet, and about 3x faster than LEN. We did not measure the runtime of MXGNet, Rel-AIR, or LEN with the teacher-model framework, but we expect them to require substantial resources and time.

\smallskip
\noindent{\bf Generalization\quad}
An important property of a model that is good at abstract reasoning is to be able to generalize over rules that were not seen during training. PGM has specially built versions of the dataset for this purpose. Aside from the `Neutral' regime, there are seven more regimes, where different rule types were omitted from the train and validation sets, and the test set consists solely of the held-out rules. For more details, please refer to Santoro et al.~\cite{santoro2018measuring}. 

Tab.~\ref{tab:generalization} shows that our method generalizes better in all regimes. Most notably are the 'HO line-type' and 'HO Triple Pair' regimes, where our method has a gap of 10\% from the baselines. We do note, however, that other regimes, such as 'HO shape-color' and 'Extrapolation', only made a negligible improvement on the baselines and are still only slightly better than random. This reassures that there is still much to be done for future work in this regard. {\color{black}In the generalization experiments, training with $\mathcal{L}_3$ only showed minor improvement. We, therefore, conclude that the performance gain for out-of-domain rules has more to do with the architecture, i.e. the multi-scale design and the relation and pattern modules.}

\smallskip
\noindent{\bf Understanding the Role of Each Scale\quad}
The premise of the importance of multi-scale design is incomplete without gaining additional knowledge on the contribution of each resolution. We, therefore, trained the model multiple times on RAVEN-FAIR with different combinations of scales ($h, m, l$) removed. We did not use weight balancing or the multi-head attentive loss for this evaluation. In Tab.~\ref{tab:multires}, we show the average accuracy per rule type.
The results clearly show that each resolution has an advantage in different rules. The lower resolution solves the 'Center', '2$\times$2', and '3$\times$3' configurations better, and the upper resolution is more adept for 'L-R', 'U-D' and 'O-IC'. The 'O-IG' configuration is best solved by the middle resolution, which suggests it requires a combination of high-resolution and low-resolution features. When using two scales, the combination of the lower and upper resolutions was better than the ones with the middle resolution, even though the middle resolution is better when only one scale is permitted. This shows that the combination of upper and lower resolutions is more informative. 
Finally, the full model performs best on average, and the added $\mathcal{L}_3$ improves on all configurations by optimizing each scale separately.

A similar analysis was done on PGM. Since PGM is a very large dataset that takes longer to train, instead of retraining, we measured the accuracy of prediction using the output of a single bottleneck by masking the outputs of two different bottlenecks each time.
Questions with multiple rules were ignored in this analysis since it wouldn't be clear which rules assisted in answering the question correctly and would contaminate the analysis. Fig.~\ref{fig:multires} visualizes the accuracy per resolution for each type of rule. As with RAVEN-FAIR, there is a clear role for each resolution. The lower resolution is responsible for the 'line' rules, which are more semantic, while the upper resolution is for the 'shape' rules, which have a strong spatial dependency. The middle resolution is specialized in shape-number, likely because the 3$\times$3 grid alignment of the shapes in each individual image $I_C^n$ of PGM (not to be confused with the 3$\times$3 question alignment) is close to the 5$\times$5 shaped matrix of the encodings $e_m$.
The analysis of the model trained on PGM\_meta shows specifically where the added auxiliary loss contributes. All resolutions improved their area of expertise from the setting without the metadata and learned to solve additional tasks. Some rules were already solved without the auxiliary loss, such as 'shape-number' and 'line-type, but others, such as 'shape-color' and 'shape-type', received a substantial increase in accuracy. 
{$\mathcal{L}_3$ showed a large improvement on all rules and better utilization of the middle resolution, without having to use the metadata.} 
The exact results on each rule can be found in the supplementary.

Both analyses come to the same conclusion: Each scale is naturally advantageous for a different set of tasks. Rules that require full embedding of the image, to recognize the shape of a large object (RAVEN 'Center') or detecting a pattern in arbitrary located lines (PGM 'line-color'), require low-resolution encoding. Rules on small objects and specific positions (RAVEN 'L-R', PGM 'shape-position') are better solved in high resolution, before the spatial coherency is lost.
We noticed during training that the model converges in steps, i.e. that several times the improvement on the validation set stagnates and then starts to rise again. We hypothesized that these steps occur when learning new rules and have conducted a per-rule analysis during training. The results, presented in the supplementary, indicate that this is indeed the case. 
A different set of rules is learned in each step and the learned rules usually have a common property.

\section{Conclusions}
The novel method we introduce outperforms the state of the art methods across tasks, protocols, benchmarks, and splits. It also generalizes better on unseen rules. Since the MC protocol can be readily exploited, we advocate for the SC protocol, which more directly tests the ability to infer the pattern from the question. For MC, evaluating on the RAVEN-FAIR variant mitigates the weaknesses of the original RAVEN benchmark. We expect both multi-scale encoders and pairwise distance-based pooling to be beneficial also for other multi-faceted tasks that involve reasoning such as causality analysis based on temporal sequences, visual question answering, and physics-based reasoning. 

\subsection*{Acknowledgements}
The contribution of the first author is part of a Ph.D. thesis research conducted at Tel Aviv University.

This project has received funding from the European Research Council (ERC) under the European Unions Horizon 2020 research and innovation programme (grant ERC CoG 725974).

\clearpage

{\small
\bibliographystyle{ieee_fullname}
\bibliography{egbib}
}

\clearpage
\appendix

\section{An example of biases in the construction of the negative choices}\label{sec:example}

In the main paper, Sec.~2.1, we discuss the importance of the selected approach for the creation of the negative choices and that it should be done carefully. In this section, we elaborate on this topic by showing examples for bad edge cases.

Consider the following four sets of answers to the question ``when and where was 'The Declaration of the Rights of Man' published?''

\begin{enumerate}
\item   \makebox[35mm]{(a) Aug. 1789, France\hfill} ~ 
        \makebox[35mm]{(b) Albert Einstein\hfill} \\
        \makebox[35mm]{(c) 55\%\hfill} ~ 
        \makebox[35mm]{(d) Alpha Centauri A\hfill}
\item   \makebox[35mm]{(a) Aug. 1789, France\hfill} ~ 
        \makebox[35mm]{(b) Nov. 1784, Italy\hfill} \\
        \makebox[35mm]{(c) Dec. 1780, Brazil\hfill} ~ 
        \makebox[35mm]{(d) Feb. 1782, Japan\hfill}
\item   \makebox[35mm]{(a) Aug. 1789, France\hfill} ~ 
        \makebox[35mm]{(b) Nov. 1789, France\hfill} \\
        \makebox[35mm]{(c) Aug. 1784, France\hfill} ~ 
        \makebox[35mm]{(d) Aug. 1789, Italy\hfill}
\item   \makebox[35mm]{(a) Aug. 1789, France\hfill} ~ 
        \makebox[35mm]{(b) Nov. 1789, France\hfill} \\
        \makebox[35mm]{(c) Nov. 1784, France\hfill} ~ 
        \makebox[35mm]{(d) Nov. 1784, Italy\hfill}
\end{enumerate}

In all options, the correct answer is the same (a). However, there is a big difference in the negative examples that make the question either easier of harder to answer.

The first option is too simple. The negative choices do not fit the domain of the question and it can be solved without any prior knowledge (by prior knowledge, we mean having some information about the questions). The only answer that fits the domain is also the correct answer. 

In the second option, all negative choices are from the same domain as the correct answer. The negative choices are similar enough so that the correct answer cannot be chosen, beyond luck, without having at least some prior knowledge. However, the options are at the same time random enough so that knowing only one attribute (month, year, location) is sufficient to know the answer. For example, one only needs to know that the declaration was published in France to correctly answer the question, without knowing when it was published.

In the third options, each negative choice is produced by changing a single attribute from the correct answer. Unlike the previous option, knowing only one attribute is not enough to eliminate all negative choices, which makes the question more difficult. However, one could notice that 'Aug', '1789' and 'France' are majority attributes in all the answers. If this pattern is recurring across all questions in the test, the participant can learn how to locate the correct answers without even looking at the question.
Paradoxically, this issue becomes more severe when more choices are given, in contrast to the purpose of adding more choices.

Finally, the fourth option overcomes all the limitations of the previous examples. The negative choices are sufficiently related so that elimination requires prior knowledge and the attributes are diverse enough so that a majority guess does not promise a correct answer. However, this option is more complicated to produce and a sophisticated method for generating the negative choices needs to be designed.

\subsection{Implications for RAVEN}

As explained in the related work (Sec.~2), the negative examples in RAVEN are generated by changing a single arbitrary attribute from the correct answer every time. This approach is equivalent to option (3) in the example given above. The problem with this approach is that the model can learn to extract the properties of each attribute from the choice images and compare which values occur in most options. The model can then make an educated guess on which choice is the correct answer.

In the next section, we show that this topic is not just a hypothetical concern, but an actual flaw of the RAVEN dataset that is easily exploited by MC protocol models.

\begin{table}[t]
\caption{Dataset benchmarks using ResNet. Accuracy(\%)}
\label{tab:resnet}
\centering
\begin{tabular}{@{}l@{~}c@{~~}c@{~}c@{~}c@{~}c@{~}c@{~}c@{}}
\toprule
& \multicolumn{3}{c}{PGM} & \multicolumn{4}{c}{RAVEN}\\
\cmidrule(lr){2-4} 
\cmidrule(lr){5-8}
Model       & all & lines & shapes & ORIG & FAIR & R-IN & R-ALL \\
\midrule
\emph{ResNet-Blind}    & 18.6 & 33.8 & 14.7 & 80.2 & 17.2 & 13.1 & 13.5 \\
\emph{ResNet-MC}      & 41.1 & 60.7 & 28.7 & 72.5 & 24.5 & 52.8 & 81.6 \\
\emph{ResNet-SC}      & 48.9 & 67.1 & 40.1 & 40.4 & 58.4 & 82.9 & 94.4 \\
\bottomrule
\end{tabular}
\end{table}

\section{Dataset analysis}\label{sec:data_analysis}

In Sec.~2.1 of the paper, we discuss the datasets and their potential pitfalls. We then define two different protocols (SC and MC) that can be followed when challenging this task. In Sec.~3, we provide a new dataset (RAVEN-FAIR) that is better suited than the original dataset for the MC protocol. We have discussed the results of experiments done on the blind, SC and MC settings. In this section, we provide the details of these experiments and their results.

Before using the PGM and RAVEN datasets in our experiments. We analyzed each of them in how they perform on three tests: (i) context-blind test, (ii) multiple choice test (iii) single choice test. The results can be seen in Tab.~\ref{tab:resnet}. The model for each test is based on the ResNet16, with an adjusted input and output layer to comply with each test. For the context-blind test, the model (ResNet-Blind) accepts the eight choices, without the context images, as an 8-channel image and produces a probability for each of the input channels to be the correct answer. For the multiple choice test, the model (ResNet-MC) accepts all 16 images as a 16-channel image (context:1-8, choices:9-16) and produces the probability for channels 9-16, which are the choices. For the single choice test, the model (ResNet-SC) accepts a 9-channel image with the context being the first eight channels and a choice added as the ninth channel. The model is fed eight iterations for each question to evaluate each choice separately, and the image with the highest score is chosen as the correct one.

For PGM, we evaluated not only the full dataset, but also two subsets of it, where rules were only applied on the lines or on the shapes. Each subset accounts for roughly a third of the dataset and the remaining third is where both line and shape rules are applied simultaneously. In Tab.~\ref{tab:resnet}, the full dataset came out to be relatively balanced, it had a low score on the context-blind test (18.6\%) and balanced score between the MC and SC protocols (41\% for MC and 48.9\% for SC). We have found the lines subset to be significantly easier overall than the shapes subset across all tests, which suggests that the dataset could be improved in that regard.

For RAVEN, we compared four different versions of the dataset. The difference between the versions is the method in which the negative choices were selected. ORIG is the original version of RAVEN. FAIR is our proposed improved dataset. R-IN and R-ALL are two versions where the negative choices were chosen randomly. In R-IN, the negative examples were sampled from the same domain as the correct answer. In R-ALL, we sampled from all domains. The four versions of the dataset represent the four options presented in the question earlier in Sec.~\ref{sec:example}. R-ALL represents option 1, since the negative choices are out of the domain of the correct answer. R-IN represents option 2, since the negative choices are randomly selected from within the domain of the correct answer. ORIG represents option 3, since the negative choices are one attribute away from the correct one. FAIR represents option 4, since it aims not to be too conditioned on the correct answer while not being too random.

The evaluation in Tab.~\ref{tab:resnet}, shows that RAVEN indeed fails the context-blind test (80.2\%). On the other hand, R-IN (13.1\%) and R-ALL (13.3\%) were unsolvable in this setting, as expected. RAVEN-FAIR passed the context-blind as well (17.2\%), with an equal performance with PGM. When the context was added, both R-IN and R-ALL became notably easy to solve (82.9\%, 94.4\% respectively on ResNet-SC), which highlights that sampling negative choices randomly is a bad design. ORIG turned out to be more difficult than FAIR in the SC test. This is reasonable since all negative examples are very close to the correct answer in ORIG, which makes them more challenging than those of FAIR. On the other hand, ORIG turned out to be really easy in the MC test (72.5\%), while FAIR was significantly harder (40.4\%). This is due to the fact that ORIG fails the context-blind test and is easy to solve when all the choices are presented simultaneously. This evaluation shows that one needs to be very careful in evaluating models with RAVEN, since it cannot be used in the MC setting.

Interestingly, except for the original RAVEN, ResNet-SC performed consistently better than ResNet-MC. This is in contrast to the common sense that MC is a simpler setting than SC due to capability to compare the choices. We conclude that this is an architectural drawback of ResNet-MC that it has less practical capability, since it processes all choices at once without added capacity. It is also not permutation invariant to the choices, which means that changing the order of the choices could lead to a different result.

\begin{table*}[t]
\caption{Understanding the role of each scale in PGM. Accuracy(\%)}
\label{tab:multires-pgm}
\centering
\begin{tabular}{@{}lllc@{~}c@{~}c@{~}cc@{~}c@{~}c@{~}cc@{~}c@{~}c@{~}c}
\toprule
& & & \multicolumn{12}{c}{Operation} \\
\cmidrule(lr){4-15}
& & & \multicolumn{4}{c}{Progression} & \multicolumn{4}{c}{XOR+OR+AND} & \multicolumn{4}{c}{Consistent Union} \\
\cmidrule(lr){4-7}
\cmidrule(lr){8-11}
\cmidrule(lr){12-15}
& Subset & Attribute & ALL & H & M & L & ALL & H & M & L & ALL & H & M & L \\
\midrule
\parbox[t]{3mm}{\multirow{7}{*}{\rotatebox[origin=c]{90}{PGM}}}
& \multirow{5}{*}{\rotatebox[origin=c]{0}{Shape}}
& Color     & 14.8 & 14.8 & 12.2 & 14.1 & 15.4 & 14.7 & 14.9 & 12.1 & 17.3 & 16.2 & 14.6 & 11.9 \\
& & Type      & 23.8 & 20.8 & 12.6 & 11.5 & 12.9 & 13.5 & 12.1 & 10.3 & 15.7 & 11.0 & 12.1 & 13.2 \\
& & Size      & 96.5 & 95.7 & 13.5 & 11.5 & 51.0 & 43.9 & 13.7 & 14.9 & 79.1 & 73.6 & 12.0 & 15.1 \\
& & Position  & \multicolumn{4}{c}{\xmark} &  99.9 & 99.9 & 12.6 & 14.7 & \multicolumn{4}{c}{\xmark} \\
& & Number    & 100 & 12.7 & 100 & 12.4 & \multicolumn{4}{c}{\xmark} & 100 & 13.7 & 100 & 13.7 \\
\cmidrule(lr){2-15}
& \multirow{2}{*}{\rotatebox[origin=c]{0}{Line}}
& Color     & 100 & 13.3 & 10.7 & 100 &  96.9 & 12.4 & 12.6 & 92.9 & 99.6 & 12.5 & 11.4 & 99.9 \\
& & Type      & \multicolumn{4}{c}{\xmark} & 100 & 39.1 & 13.1 & 30.4  & 100 & 100 & 15.9 & 12.8 \\
\midrule
\parbox[t]{3mm}{\multirow{7}{*}{\rotatebox[origin=c]{90}{PGM\_meta}}}
& \multirow{5}{*}{\rotatebox[origin=c]{0}{Shape}}
& Color     & 71.5 & 58.9 & 11.8 & 15.6 & 31.4 & 29.0 & 12.3 & 12.0 & 40.8 & 28.5 & 12.7 & 15.4 \\
& & Type      & 92.6 & 88.8 & 13.0 & 17.5 & 52.0 & 47.8 & 15.2 & 13.5 & 65.8 & 47.7 & 12.8 & 12.8 \\
& & Size      & 92.3 & 91.2 & 12.7 & 15.0 & 56.3 & 45.6 & 11.7 & 12.1 & 72.5 & 65.9 & 15.1 & 14.3 \\
& & Position  & \multicolumn{4}{c}{\xmark} &  99.9 & 94.6 & 68.8 & 12.5 & \multicolumn{4}{c}{\xmark} \\
& & Number    & 100 & 14.8 & 100 & 12.1 & \multicolumn{4}{c}{\xmark} & 99.6 & 13.7 & 99.4 & 14.6 \\
\cmidrule(lr){2-15}
& \multirow{2}{*}{\rotatebox[origin=c]{0}{Line}}
& Color     & 100 & 12.3 & 13.8 & 100 &  96.0 & 11.3 & 12.4 & 87.3 & 100 & 12.1 & 10.6 & 99.9 \\
& & Type    & \multicolumn{4}{c}{\xmark} & 100 & 50.8 & 12.9 & 49.2  & 100 & 99.9 & 15.7 & 23.4 \\
\midrule
\parbox[t]{3mm}{\multirow{7}{*}{\rotatebox[origin=c]{90}{PGM $\mathcal{L}_3$}}}
& \multirow{5}{*}{\rotatebox[origin=c]{0}{Shape}}
& Color     & 91.5 & 89.5 & 12.7 & 13.7 & 54.5 & 48.7 & 12.0 & 12.7 & 88.0 & 83.3 & 12.2 & 13.1 \\
& & Type      & 98.2 & 97.2 & 14.1 & 12.1 & 80.0 & 76.9 & 12.6 & 12.5 & 91.3 & 88.5 & 13.0 & 11.5 \\
& & Size      & 98.0 & 92.5 & 11.7 & 12.6 & 79.8 & 76.4 & 12.8 & 12.4 & 93.9 & 88.2 & 12.6 & 13.9 \\
& & Position  & \multicolumn{4}{c}{\xmark} &  99.9 & 98.5 & 99.9 & 12.5 & \multicolumn{4}{c}{\xmark} \\
& & Number    & 100 & 27.2 & 100 & 12.1 & \multicolumn{4}{c}{\xmark} & 100 & 25.3 & 99.9 & 12.5 \\
\cmidrule(lr){2-15}
& \multirow{2}{*}{\rotatebox[origin=c]{0}{Line}}
& Color     & 100 & 12.6 & 99.2 & 100 &  97.7 & 14.6 & 87.7 & 97.2 & 100 & 12.5 & 97.9 & 98.3 \\
& & Type    & \multicolumn{4}{c}{\xmark} & 99.9 & 99.5 & 99.8 & 39.3  & 100 & 99.6 & 99.8 & 34.4 \\
\bottomrule
\end{tabular}
\end{table*}

\begin{figure*}[t]
\centering
    \includegraphics[width=\textwidth, trim={70 10 70 20}, clip]{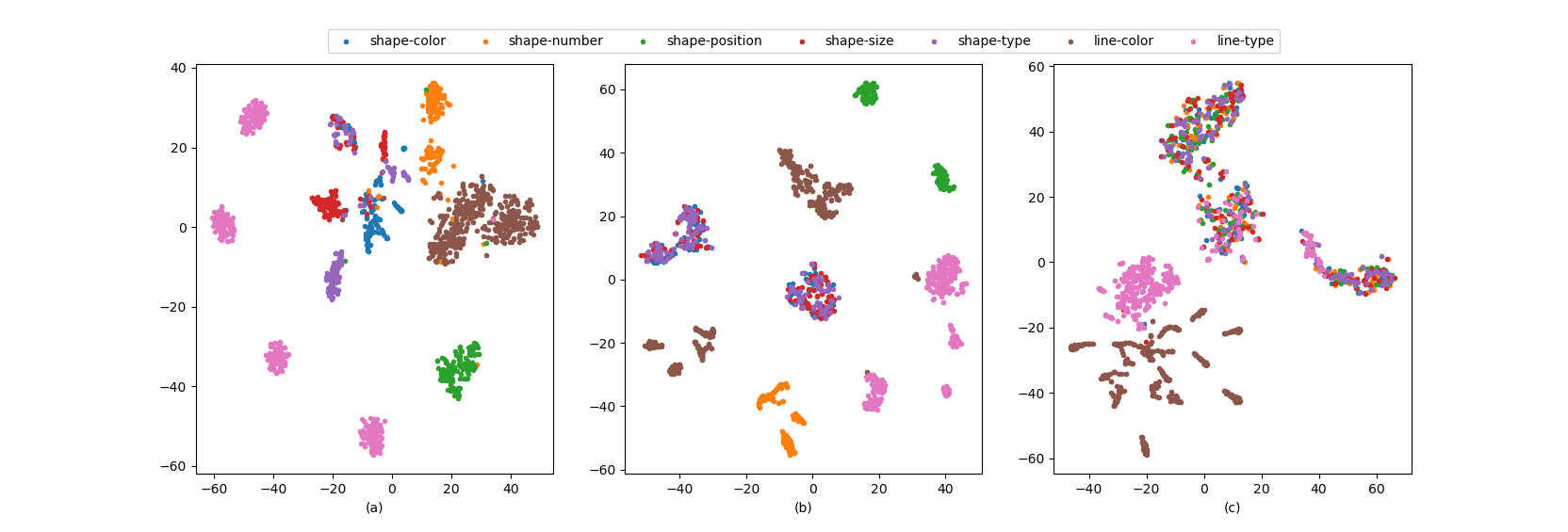}
    \caption{t-SNE analysis of each resolution. (a) High resolution $v_h$. (b) Middle resolution $v_m$. (c) Low resolution $v_l$.}
    \label{fig:tsne}
\end{figure*}

\begin{figure*}[t]
\centering
    \begin{tabular}{cc}
    \includegraphics[width=0.45\textwidth, trim={0 0 0 0}, clip]{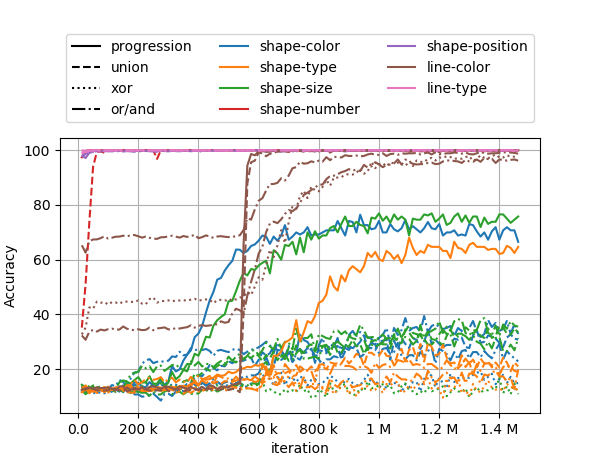} &
    \includegraphics[width=0.45\textwidth, trim={0 0 0 0}, clip]{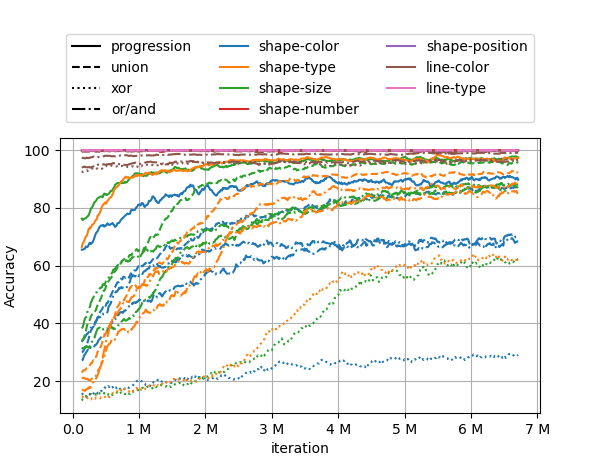} \\
    (a) & (b) \\
    \end{tabular}
    \caption{Accuracy for each rule over time. (a) Without $\mathcal{L}_3$. (b) With $\mathcal{L}_3$. It can be seen that during each 'step', the model is focused at learning a different subset of tasks. It is also noticeable that the added multihead loss immediately improves the rules that the model was not able to solve.}
    \label{fig:accuracy_over_time}
\end{figure*}

\section{Analyzing PGM per rule and stage}

In Sec.~5, we analyze the performance on PGM per rule and per stage in our networks. This allows us to determine which stage is responsible for solving each task. We have shown visually how each stage performs on each task. In addition to that, Tab.~\ref{tab:multires-pgm} shows the accuracy for in total and for each stage with respect to each rule. The table also shows the accuracy per type of operation ('progression', 'logical', 'consistent-union'), where this separation was not done in the main paper. We noticed that the operations have sometimes different performances. For example, in PGM\_meta,  the rules 'shape-color', shape-type' and 'shape-size' work much better in 'progression' than in the other operations. We also noticed that, for each rule, the same resolution solves all operations.

To better understand how each resolution solves each type of rule, we performed a t-SNE analysis of the embedding layer $v_t$ of each resolution, for the embedding of the correct answer $I^{a^*}$. The analysis was done on the fully trained model with $\mathcal{L}_3$. It can be seen in Fig.~\ref{fig:tsne}. It should be noted that separation of the rules in the latent space is not a requirement for accuracy in solving the task, but we have observed high correlation between the two in each resolution and it is a valuable analysis nevertheless.
(a) The upper resolution is good at separating most rule types. It even separates 'line-type' into distinct sub-groups, which we have found to be correlated to the relational operation 'progression', 'union', 'xor', etc'.
(b) The middle resolution is very good at separating some tasks, such as 'line-type', 'line-color', 'shape-number', and 'shape-position'. These rules are also separated into sub-groups and are also where this resolution has a high accuracy on. The other rules are not separated at all and the resolution also had trouble in solving them. (c) The lower resolution had the lowest overall performance in accuracy and also in separating the rules. However, we did notice that it was exceptional in separating the 'line-color' rules into multiple sub-groups. The lower resolution also has the highest accuracy on this type of rules.

Aside from the final results per rule, we noticed an interesting behaviour in the training convergence. Fig.~\ref{fig:accuracy_over_time} shows the accuracy of each rule and operation over time. We noticed that the model does not learn all the rules at the same time. Instead, the training appears to go in stages, where the model learns a particular kind of rule at each stage. 
In Fig.~\ref{fig:accuracy_over_time}(a), we show the learning progress without $\mathcal{L}_3$, and in Fig.~\ref{fig:accuracy_over_time}(b), we show the contribution of $\mathcal{L}_3$ when it is added. It was especially surprising to see that the model only started to learn 'line-color' very late (around 500K iterations) but did it very fast (within ~50K iterations for 'progression' and 'union'). The model is able to learn numerous rules without the multihead loss $\mathcal{L}_3$. However, many rules, especially in the 'shape' category are not solved well. The addition of the multihead loss immediately and significantly increases the performance on all rules.

Since training took a long time and we measured only after each epoch, the plot doesn't show the progress in the early stages of the training, which would show when the 'easier' tasks have been learned. Future work can focus on this 'over-time' analysis and try to explain: (i) 'why are some rules learned and others not?', (ii) 'why are some rules learned faster than others?', (iii) 'how can the training time be shortened?', (iv) 'would the rules that were learned late also be learned if the easier rules were not present?'.

\section{Architecture of each sub-module}

We detail each sub-module used in our method in Tab.~\ref{table:arch_E}-\ref{table:arch_MLP}. Since some modules re-use the same blocks, Tab.~\ref{table:arch_general} details a set of general modules.

\begin{table*}[t]
\caption{General modules, with variable number of channels $c$.}\label{table:arch_general}
\centering
\begin{tabular}{llccc}
\toprule
Module  & layers & parameters & input & output\\ 
\midrule
\multirow{7}{*}{ResBlock3($c$)}  
        & Conv2D & C$c$K3S1P1 & $x$ \\
        & BatchNorm \\
        & ReLU \\
        & Conv2D & C$c$K3S1P1 \\
        & BatchNorm & & & $x'$ \\
        & Residual & & $(x,x')$ & $x''=x+x'$ \\
        & ReLU \\
\midrule
\multirow{10}{*}{DResBlock3($c$)}
        & Conv2D & C$c$K3S1P1 & $x$ \\
        & BatchNorm \\
        & ReLU \\
        & Conv2D & C$c$K3S1P1 \\
        & BatchNorm & & & $x'$ \\
        \cmidrule(lr){2-5}
        & Conv2D & C$c$K1S2P0 & $x$ \\
        & BatchNorm & & & $x_d$ \\
        \cmidrule(lr){2-5}
        & Residual & & $(x_d,x')$ & $x''=x_d+x'$ \\
        & ReLU \\
\midrule
\multirow{7}{*}{ResBlock1($c$)}
        & Conv2D & C$c$K1S1P0 & $x$ \\
        & BatchNorm \\
        & ReLU \\
        & Conv2D & C$c$K1S1P0 \\
        & BatchNorm & & & $x'$ \\
        & Residual & & $(x,x')$ & $x''=x+x'$ \\
        & ReLU \\
\bottomrule
\end{tabular}
\end{table*}

\begin{table*}[ht]
\caption{Encoders $E_h,E_m,E_l$}\label{table:arch_E}
\centering
\begin{tabular}{lllcc}
\toprule
Module  & layers & parameters & input & output\\ 
\midrule
\multirow{6}{*}{$E_{h}$}       
        & Conv2D    & C32K7S2P3 & $I^1$ & \\
        & BatchNorm \\
        & ReLU \\
        & Conv2D    & C64K3S2P1 \\
        & BatchNorm \\
        & ReLU & & & $e^1_h$ \\
\midrule
\multirow{6}{*}{$E_{m}$}       
        & Conv2D    & C64K3S2P1 & $e^1_h$ & \\
        & BatchNorm \\
        & ReLU \\
        & Conv2D    & C128K3S2P1 \\
        & BatchNorm \\
        & ReLU & & & $e^1_m$ \\
\midrule
\multirow{6}{*}{$E_{l}$}       
        & Conv2D    & C128K3S2P1 & $e^1_m$ & \\
        & BatchNorm \\
        & ReLU \\
        & Conv2D    & C256K3S2P1 \\
        & BatchNorm \\
        & ReLU & & & $e^1_l$ \\
\bottomrule
\end{tabular}
\end{table*}

\begin{table*}[ht]
\caption{Relation networks $RN_h,RN_m,RN_l$}\label{table:arch_RM}
\centering
\begin{tabular}{lllcc}
\toprule
Module  & layers & parameters & input & output \\
\midrule
\multirow{5}{*}{$RN_{h}$}       
        & Conv2D    & C64K3S1P1 & $(e^1_h,e^2_h,e^3_h)$ & \\
        & ResBlock3 & C64 \\
        & ResBlock3 & C64 \\
        & Conv2D    & C64K3S1P1 \\
        & BatchNorm & & & $r^1_h$ \\
\midrule
\multirow{5}{*}{$RN_{m}$}       
        & Conv2D    & C128K3S1P1 & $(e^1_m,e^2_m,e^3_m)$ & \\
        & ResBlock3 & C128 \\
        & ResBlock3 & C128 \\
        & Conv2D    & C128K3S1P1 \\
        & BatchNorm & & & $r^1_m$ \\
\midrule
\multirow{5}{*}{$RN_{l}$}       
        & Conv2D    & C256K1S1P0 & $(e^1_l,e^2_l,e^3_l)$ & \\
        & ResBlock1 & C256 \\
        & ResBlock1 & C256 \\
        & Conv2D    & C256K1S1P0 \\
        & BatchNorm & & & $r^1_l$ \\
\bottomrule
\end{tabular}
\end{table*}

\begin{table*}[ht]
\caption{Bottlenecks $B_h,B_m,B_l$}\label{table:arch_B}
\centering
\begin{tabular}{lllcc}
\toprule
Module  & layers & parameters & input & output \\
\midrule
\multirow{2}{*}{$B_{h}$}       
        & DResBlock3 & C128 & $b_h$ \\
        & DResBlock3 & C128 & & \\
        & AvgPool2D & & & $v_h$ \\
\midrule
\multirow{2}{*}{$B_{m}$}       
        & DResBlock3 & 256 & $b_m$ \\
        & DResBlock3 & C128 & & \\
        & AvgPool2D & & & $v_m$ \\
\midrule
\multirow{4}{*}{$B_{l}$}       
        & Conv2D    & C256K1S1P0 & $b_l$ & \\
        & BatchNorm \\
        & ReLU \\
        & ResBlock1 & C128 & & $v_l$ \\
\bottomrule
\end{tabular}
\end{table*}

\begin{table*}[ht]
\caption{MLP}\label{table:arch_MLP}
\centering
\begin{tabular}{lllcc}
\toprule
Module  & layers & parameters & input & output \\
\midrule
\multirow{8}{*}{$MLP$}       
        & Linear & C256 & $(v_h,v_m,v_l)$ \\
        & BatchNorm \\
        & ReLU \\
        & Linear & C128 & \\
        & BatchNorm \\
        & ReLU \\
        & Linear & C1 & \\
        & Sigmoid & & & $p(y=1 | I^a, I_C)$ \\
\bottomrule
\end{tabular}
\end{table*}

\end{document}